\pdfoutput=1

\documentclass[11pt]{article}

\usepackage[]{EMNLP2022}

\usepackage{times}
\usepackage{latexsym}

\usepackage[T1]{fontenc}

\usepackage[utf8]{inputenc}

\usepackage{microtype}

\usepackage{inconsolata}

\usepackage{amsmath}
\usepackage{amsfonts}
\usepackage{mathtools}
\usepackage{enumitem}
\usepackage{booktabs}
\usepackage{graphicx}
\usepackage{color}
\usepackage{comment}
\usepackage{multirow}
\usepackage{siunitx}
\usepackage{subcaption}
\usepackage{booktabs}
\usepackage{multirow}
\usepackage{amssymb}
\usepackage{calrsfs}
\DeclareMathAlphabet{\pazocal}{OMS}{zplm}{m}{n}

\title{Towards Pragmatic Production Strategies\\ for Natural Language Generation Tasks}

\author{Mario Giulianelli \\
  Institute for Logic, Language and Computation \\
  University of Amsterdam \\
  \texttt{m.giulianelli@uva.nl}}

\begin{document}
\maketitle

\begin{abstract}
This position paper proposes a conceptual framework for the design of Natural Language Generation (NLG) systems that follow efficient and effective production strategies in order to achieve complex communicative goals. In this general framework, efficiency is characterised as the parsimonious regulation of production and comprehension costs while effectiveness is measured with respect to task-oriented and contextually grounded communicative goals.
We provide concrete suggestions for the estimation of goals, costs, and utility via modern statistical methods, demonstrating applications of our framework to the classic pragmatic task of visually grounded referential games and to abstractive text summarisation, two popular generation tasks with real-world applications. In sum, we advocate for the development of NLG systems that learn to make pragmatic production decisions from experience, by reasoning about goals, costs, and utility in a human-like way. \looseness=-1
\end{abstract}

\section{Introduction}
\label{sec:intro}

Novelists choose the right words to keep readers engaged and enthused, good journalists can convey facts clearly and convincingly, while poets may want to surprise the reader. Teachers adapt their explanations to the level of their students, and the language of parents changes with the proficiency of their children, with the same objects described first using simplified funny expressions (\textit{`moo moo'}) and then more informative and discriminative names (\textit{`cow'}, \textit{`calf'}).
Using language to communicate successfully requires effort.
On the side of the language producer, it is first of all effortful to come up with words that truthfully correspond to one's communicative intent.
Then, words must be actually produced, e.g.\ said out loud or typed on a keyboard. 
At the same time, the producer has to take into consideration whether the comprehender---for whom, too, linguistic communication is costly---will be able to infer the original intent.
Comprehenders make efforts to pay attention to the utterance they are being addressed with, to interpret it, and to infer their interlocutor's communicative intent.
Fortunately, these efforts are often not in vain. They allow people to exchange knowledge, ideas, plans, and to achieve goals.

This paper introduces a conceptual framework for Natural Language Generation (NLG) in variably complex communicative scenarios, which relies on three main notions: \textbf{communicative goals}, production and comprehension \textbf{costs}, and \textbf{utility}. We define these notions formally and then, in two case studies, we provide suggestions for their operationalisation in classic NLG tasks. In sum, we model humans as decision makers striving for efficient and effective communication, and argue that human-like linguistic behaviour emerges as a result of reasoning about goals, costs, and utility. Learning to navigate the complex decision space defined by these notions is still an open problem: we discuss possible promising directions.

\section{Doing Things with Words}
\label{sec:definitions}
Communication always comes with a goal: speakers use words to change the state of the world. In this section, we give a characterisation of communicative goals, discuss the types of effort (or costs) necessary to achieve goals, and describe the rewards associated with successful communication.

\subsection{Communicative Goal}
\label{sec:definitions-goal}
What do speakers do with words? The communicative goal (or \textit{communicative intent}) of a speaker can be formulated as a function of the current state of the world $w \in W$: 
\begin{align}
    G_{s} \colon\ W \to W,\ \ w \mapsto w^{*}
    \label{eq:comm-goal}
\end{align}
where $w^{*}$ is the intended future state of the world. Speaker $s$ and audience $a$ are included in $w$ as they can be both conceptualised, and there is evidence that they are processed \cite{brown2015people}, as parts of the state of the world. For communication to be successful, the audience must be able to reconstruct the original communicative goal: their \textit{decoded} transformation of the world, $D_{a}\colon\!W\!\to\!W$, must be such that $D_{a}(w)\!\approx\!G_{s}(w)$.\footnote{We sometimes refer to $D_{a}(w)$ and $G_{s}(w)$ as $D_{a}$ and $G_{s}$, as in our formulation these functions are always applied to the current state of the world $w$.}

Communicative goals shape and constrain a speaker's production choices: different utterance types typically correspond to different goals.
The communicative goal of a referring utterance (\textit{``The black and white cat''}), for example, is a state of the world where the audience is able to identify an entity in context. The transformation $D_{a}$ required to achieve $w^{*}$ is a change of attention by the audience. 
Statements (\textit{``The Sun is a star''}) are typically used when the purpose of an interaction is pure information transmission---e.g., when giving a scientific talk. In this case, the communicative goal is a state of the world in which the audience holds new beliefs, the ones intended by the speaker. $D_{a}$~is a transformation of the belief state of the audience, and the communicative goal is achieved when $D_{a}(w)\!\approx\!G_{s}(w)\!=\!w^{*}$.
All utterance types---e.g., questions, directives, and performatives---can be seen as strategies to achieve communicative goals. 
The same utterance type, and even the same utterance, can fulfil different goals: a blatantly false statement (\textit{``It never rains in Amsterdam''}) can be used for comedic effect rather than for conveying facts. For simplicity, in the rest of this paper, we describe utterances as having a single communicative goal. Often, however, different goals are associated with the same utterance at the same time: a teacher can use a question (\textit{``Are you sure this is the right answer?''}) to inform their student that their answer is incorrect, while showing a positive attitude towards them---thereby striving for both epistemic and social utility. Our framework naturally generalises over such cases; when multiple communicative goals are involved, states of the world can be designed accordingly.\footnote{
To account for epistemic and social utility, for example, states of the world can be defined to include the audience belief state as well as their emotional state.
}\looseness-1

\subsection{Production Costs}
\label{sec:definitions-production}
Given the current and the intended future state of the world, $w$ and $w*\!=\!G_s(w)$, a speaker \textit{encodes} the communicative goal $G_{s}(w)$ into a mental representation of the intended state of the world: $E_{s}(G_{s}(w))\!=\!e$. To use a slightly different vocabulary, this is the speaker's \textit{conception} of the intended environment state. The speaker then \textit{realises} $e$ as an utterance $r$ which is presented to the audience: $R_s(e)\!=\!r$. Two types of cost are associated with the encoding and realisation processes. Because the encoding process is inevitably lossy---mental representations are compressed representations of the real state of the world---the speaker makes an effort to reduce information loss; we refer to this as the \textbf{encoding cost} $C^E$. 
The cost associated with executing a bit of behaviour $r$ meant to be perceived by the audience (e.g., speaking, writing, or typing) is the \textbf{realisation cost} $C^R$. Both costs affect the decision making process of speakers. In addition, the speaker is influenced by the expected comprehension costs of the audience.

\subsection{Comprehension Costs}
\label{sec:definitions-comprehenion}
The speaker's communicative goal $G_{s}(w)$ is not observable by comprehenders. Given a state of the world $w$ and the speaker's behaviour $r$, comprehenders \textit{process} $r$ into a reconstruction of the original mental representation, $P_a(r)\!=\!e'\!\approx\!e$, from which they \textit{decode} the speaker's communicative goal: $D_a(e')\!=\!w'\!\approx\!G_{s}(w)$. Two types of cost are associated with the comprehension of an utterance. Speaker and comprehender are different individuals and therefore have different ways of encoding communicative goals into messages \cite{ConnelLynott2014}. In the absence of a perfect model of the speaker's encoding mechanism, reconstructing $e$ is a lossy and effortful process; we denote the corresponding cost as \textbf{processing cost},~$C^P$. The second cost results from interpreting $e'$ in context---i.e., decoding from $e$ the state of the world intended by the speaker. In other words, this is the effort required to \textit{ground} the message in the environment. We refer to it as the \textbf{decoding cost}~$C^D$. It is important to note that although processing and decoding costs are on the side of comprehenders, speakers estimate them and take them into account when making production decisions.

\subsection{Utility}
\label{sec:definitions-utility}
In what ways is the decision making process of speakers affected by these costs? Speakers are thought to be driven by efficiency concerns \cite{zipf1949human,jaeger2011language}: they strive to minimise the collaborative effort required to achieve their communicative goals \cite{clark1986referring,clark1989contributing}. We thus take the speaker's utility $U_s$ to be inversely proportional to the joint production and comprehension costs required for goal achievement~($D_{a}\!\approx\!G_{s}$). Production costs can be reduced directly by the speaker, by putting less cognitive and physical effort in encoding and realisation. Comprehension costs, instead, need to be first estimated via a mental model of the audience's comprehension system (including their conceptual knowledge, perceptual capacity, language proficiency, etc.). The ability to form such mental models is often referred to as Theory of Mind \cite{premack1978tom} and it is deemed a fundamental social-cognitive skill for language acquisition and language use \cite{tomasello2005constructing}.\looseness-1

Speaker's utility is not only defined in terms of costs; speakers profit from getting things done with their words. Thus $U_s$ is directly proportional to the positive cognitive, physical, and social effects that derive from achieving the intended state of the world $w^{*}$. Because, in practice, interlocutors often approach but do not reach $w^{*}$ exactly, $U_s$ can be defined as a function of $D_a(w)$ and $G_s(w)$ that quantifies the difference in positive effects between true and intended states of the world.

\section{Case Study 1: Reference Games}
\label{sec:case-study-1}
In this section, we demonstrate how to use our framework to conceptualise a communication scenario that corresponds to a classic NLG task, referring expression generation \citep{reiter1997building,krahmer-van-deemter-2012-computational}. 
We will also provide concrete examples of how to model the costs and utility described in Section~\ref{sec:definitions}.

In a reference game, the goal is for participants to produce descriptions that allow comprehenders to identify the correct referent out of a set of candidates. These games have been extensively used in psycholinguistics to study human strategies for effective reference \cite[][]{krauss1964changes,BrennanClark1996,hawkins2020characterizing}.  
For our case study, we use a visually grounded reference game with two participants, a speaker $s$ and a listener $a$. The speaker produces referring utterances $r$ such as \textit{``a boy cutting a cake''} and the listener needs to identify the target image $i^{*}$ among a set of similar images $V$, the visual context \cite[see, e.g.,][]{shore-etal-2018-kth,haber2019photobook}.
The initial state of the world is one where the speaker is aware of the target referent while the listener has no information about it. We can express such a state of the world as $w = (V,p_s,p_a)$, i.e. in terms of the speaker and listener's probability distributions $p_s$ and $p_a$ over candidate images $V$ before anything is uttered ($r\!=\!\epsilon$, the empty string):\footnote{
This setup corresponds to one-shot reference games. In multi-turn dialogues, $w$ should also include the game history.\looseness-1
}
\begin{align}
    p_s(I|V) &: p_s(I=i^*|V) = 1\\
    p_a(I|V,\epsilon) &: p_a(I=i|V,\epsilon) = \frac{1}{|V|} \ \ \forall i \in V
\end{align}
Note that $p_s$ is never observable by $a$, and for this scenario to be realistic, $p_a$ should also not be observable by $s$.
The communicative goal $G_s$ is a transformation of $w$ into $w^*$, a state of the world in which $a$ identifies $i^{*}$ as the target referent:
\begin{align}
    G_s(w) =\ &(V,p_s,p'_a)\  \text{with}\\
    &p'_a(I=i^*|V,r) = 1
    \label{eq:listener-correct}
\end{align}

How can the costs associated with reaching this state of the world using utterance $r$ be estimated?
A computer vision model may be used to encode the communicative goal $w^* = (V,p_s,p'_a)$ into a mental representation. This model receives as input the visual context $V$ and information about the target image $p_s$ and yields a mental (abstract) representation~$e\!=\!E_s(w^*)$. If this is, e.g., a model that produces image segmentations, the encoding effort $C^E$ can be quantified as the uncertainty of the model over its segmentation decisions, as the number of output image segments, or, if the segments form a scene graph, as a measure of the graph complexity.
The encoding $e$ may then be fed to an NLG model $R_s$ which \textit{realises} it into an utterance $r\!=\!R_s(e)$. The realisation cost~$C^R$ can be computed as the utterance length, the depth of the syntactic tree corresponding to the utterance, or as a function of the distribution of vocabulary ranks for the sampled utterance tokens.\looseness-1

Next, $r$ is received by the listener, who \textit{processes} it into a reconstruction of the original mental representation: $e'\!=\!P_a(r)$. This can be achieved using a neural language model, the processing cost~$C^P$ being calculated as the model's cumulative surprisal (the sum of the per-word information content). From $e'$ the listener decodes a state of the world $w'$. The decoding system may be one that measures the similarity of $e'$ to candidate image embeddings and outputs a probability distribution over $V$. The decoding cost $C^D$ can be estimated as the entropy reduction with respect to the prior probability $p_a(I|V)$ (the information gain), or as the increase in the target image's probability. 
Communication is successful if $p'_a(I=i^*|V,r) = 1$ (see Eq.\ \ref{eq:listener-correct}); in practice the condition is often relaxed to:
\begin{align}
    i^* = \arg\max_{i \in V} p'_a(I=i|V,r)
    \label{eq:listener-argmax}
\end{align}
In a simplified reference game where $p_a$ is observable by $s$, the speaker's positive utility $U_s$ can be simply modelled as $\log p'_a(i^*|V,r) - \log p_a(i^*|V)$. In a more realistic scenario, either the speaker entertains a mental model of $p_a$ and uses it to compute utility, or the listener must in turn execute a bit of behaviour to communicate the state of $p'_a$, for example by selecting an image through a simple decision rule (e.g., $\arg\max p'_a$).
$U_s$ can then be modelled as a binary reward based on the listener's behaviour: 1 for a correct guess, 0 for an incorrect one.
Recall that $U_s$ is not only a function of positive cognitive effects. It is also inversely proportional to the costs $C^E, C^R, C^P,$ and $C^D$.\looseness-1

\section{Case Study 2: Text Summarisation}
\label{sec:case-study-2}
With our second case study, we demonstrate the generality of our framework by applying it to text summarisation, a widely studied NLG (and NLU) task with a large range of practical applications. 
When people summarise a text, they produce a concise and meaning-preserving version of that text with the goal of conveying to the audience the text's most important ideas.
In NLP, texts have been typically summarised either via extraction of their most significant sentences \cite{luhn1958automatic,edmundson1969new} or by the generation of fewer, new sentences \cite{dejong1982overview,banko2000headline}. Here, we look at the second case, often referred to as \textit{abstractive summarisation}, where a summariser~$s$ produces an utterance~$r$ made up of one or multiple sentences to succinctly report the main content of a text~$t$ to an audience~$a$.
The initial state of the world is one where the summariser knows the content of $t$ while the audience has no information about it.

Summaries can have multiple communicative goals---sometimes simultaneously---roughly corresponding to practical goals of NLP summarisation systems. For example, the communicative goal $G_s$ of a summary can be a transformation of the state of the world into one in which $a$ knows the general topic of $t$ and is interested in reading~$t$. This setup roughly corresponds to headline generation, a classic abstractive summarisation task. If the practical goal of the summary, instead, is to make the audience aware of the main facts reported in a text, the communicative goal $G_s$ is a transformation of the state of the world into one in which those facts are part of $a$'s knowledge. This is the goal, for example, of summaries of financial, legal, or medical reports. \looseness-1

We now look at this second case, providing examples of how to model communicative goals, costs, and utility.
A hierarchical language model with explicit attention over multiple sentences can be used to encode the document into a mental representation $e$. The encoding cost $C^E$ can be quantified as the entropy of the attention distribution---the rationale being that it is harder to condense the information in a document in which each sentence contains salient details.
The encoding $e$ may then be fed to a generation model $R_s$ which realises it into an utterance $r\!=\!R_s(e)$ (one or multiple sentences). The realisation cost $C^R$ can be computed as the utterance length or as a function of the predicted tokens' probabilities.
The summary $r$ is received by the audience, for example via a neural language model pretrained on summaries, which processes it into a reconstruction of the original mental representation: $e'\!=\!P_a(r)$. The processing cost $C^P$ can be calculated as the model’s cumulative surprisal.
From $e’$, the audience decodes a new state of the world, one where it can hopefully answer factual questions about the target document correctly.
The decoding system can be a question answering model (which can be as simple as a table-lookup and as complex as a response generation model) and the decoding cost $C^D$ can be estimated as the system's reduction in uncertainty in answering a set of questions designed to probe understanding of the main content of the document---formulated, e.g., as key-value queries or using natural language.
The speaker’s utility $U_s$ can be modelled as the accuracy of the audience in answering questions about the content of the document.\looseness-1

\section{Pragmatic Production Strategies}
\label{sec:discussion}
Language producers are thought to balance their own production costs and their audience's comprehension costs in a way that minimises joint collaborative effort \cite{clark1986referring,clark1989contributing} while attempting to gain utility from successful communication. Nevertheless, most modern NLG systems, whose aim is arguably to reproduce the communicative behaviour of human language users, do not take into consideration the costs and utility for which humans are constantly optimising. As a major example, GPT-3 \cite{gpt3}, one of the best foundation models currently available for NLG, conflates all costs into a single next-word probability value. To generate words from this model, typically, next-word probabilities are passed to a decoding algorithm such as beam search or nucleus sampling \cite{reddy-1977,holtzman2019curious}. This algorithm can be seen as a way to search through the space of possible utterances by following a simple utility-maximising decision rule, with higher probability utterances having higher utility.
Future work should investigate decision making rules that take into account production and comprehension costs more explicitly, connecting them to the goal of the linguistic interaction. The Rational Speech Act model \cite[RSA;][]{frank2012predicting} is a compelling solution: it was shown to optimise the trade-off between expected utility and communicative effort and it is related to Rate-Distortion theory \cite{shannon1948}, the branch of information theory that studies the effect of limited transmission resources on communicative success \cite{zaslavsky-etal-2021-rate}. Its application to simple reference games has indeed demonstrated that richer decision making routines, grounded in listeners' actions and beliefs, result in human-like pragmatic behaviour \cite{sumers2021extending}. Bounded rationality \cite{simon1990bounded}, which models optimal decision making under constrained cognitive resources, is a strong alternative to RSA theory but there is so far only limited evidence that it can be used to characterise language production choices \cite{franke2010vagueness}. A third, more practically oriented solution, are utility-based decoding algorithms---e.g., minimum Bayes risk \cite{goel2000minimum} decoding---which have been successfully used to weigh in utilities and costs when selecting utterances for NLG tasks \cite{kumar2002minimum,kumar2004minimum}.

Modelling and artificially reproducing human communicative behaviour requires advanced decision making algorithms that are able to \textit{learn from experience} efficient and effective strategies for weighing costs and utility. The learned strategies should apply both to individual utterances and to sequences of utterances: this will allow successful multi-turn planning of communicative subgoals and strategies. Reinforcement learning~(RL) can naturally interact with notions of cost and utility (these can be used as learning signal for RL models, or they can be inferred by RL models from observations of human behaviour) and it has been used in combination with RSA and bounded rationality; it thus appears to be a promising avenue for the strategy learning problem. \looseness-1

Independent of the choice of language model---which is an important open question---we believe that our conceptual framework can account for a variety of human behavioural patterns of communication as described in pragmatics, the field of linguistics which studies the aspects of language use that involve reasoning about context, goals and beliefs. Let us take as an emblematic example Grice's four maxims of conversation \cite{grice1975logic}.
The maxim of \textit{quantity}, which states that speakers should make their contribution as informative as required for the current purposes of the exchange, can be understood as the optimisation of realisation and processing costs, $C^R$ and $C^P$, while ensuring that the distance from the communicative goal is reduced. The maxim of \textit{quality}, which is about making truthful contributions, can be thought of as the result of minimising decoding cost $C^D$ and maximising the probability of achieving the communicative goal.
The maxim of \textit{relation}, stating that speakers should provide information that is relevant to the exchange, can be seen as a way to ensure that production and comprehension costs are always balanced by gains in positive utility.
Finally, the maxim of \textit{manner} states that speakers should avoid obscurity of expression, ambiguity, and strive for brief and orderly contributions. This can be easily understood as the optimisation of realisation and processing cost, $C^R$ and $C^P$, given fixed encoding and decoding costs $C^E$ and $C^D$. 

\section{Conclusion}
\label{sec:conclusion}
We have presented a conceptual framework for natural language generation that relies on three central notions: communicative goals, production and comprehension costs, and utility optimisation. We have defined these notions formally and demonstrated their application to two realistic communication scenarios, providing examples for the modelling of goals, costs, and utility with modern method of statistical learning. We have further argued for our framework's ability to account for a variety of pragmatic patterns of communicative behaviour, highlighting the importance of the development of new complex decision making algorithms that learn to reproduce human-like production strategies from experience.\looseness-1
\section*{Limitations}
\label{sec:limitations}
Some of the notions upon which our framework relies are not new; they are the results of decades of research in linguistics, cognitive science, and psychology, as acknowledged in the paper. We want to highlight this here in fairness, yet we believe that there is value in bringing ideas together from a pool of interdisciplinary studies and organising them into a structured framework.
Moreover, although our proposal is designed to model language production in varying communicative scenarios, we presented only two case studies. We plan to demonstrate the generality of our framework with further case studies accompanied by computational experiments, which are absent in this paper.

\section*{Acknowledgements}
I would like to thank Raquel Fern\'andez, Arabella Sinclair, and the members of the Dialogue Modelling Group of the University of Amsterdam for our insightful discussions, as well as the anonymous EMNLP 2022 reviewers for their useful comments. This project has received funding from the European Research Council (ERC) under the European Union's Horizon 2020 research and innovation programme (grant agreement No.\ 819455).

\bibliography{anthology,custom}

\begin{thebibliography}{32}
\expandafter\ifx\csname natexlab\endcsname\relax\def\natexlab#1{#1}\fi

\bibitem[{Banko et~al.(2000)Banko, Mittal, and Witbrock}]{banko2000headline}
Michele Banko, Vibhu~O Mittal, and Michael~J Witbrock. 2000.
\newblock Headline generation based on statistical translation.
\newblock In \emph{Proceedings of the 38th Annual Meeting of the Association
  for Computational Linguistics}, pages 318--325.

\bibitem[{Brennan and Clark(1996)}]{BrennanClark1996}
Susan~E. Brennan and Herbert~H. Clark. 1996.
\newblock Conceptual pacts and lexical choice in conversation.
\newblock \emph{Journal of Experimental Psychology: Learning, Memory, and
  Cognition}, 22:1482--1493.

\bibitem[{Brown et~al.(2020)Brown, Mann, Ryder, Subbiah, Kaplan, Dhariwal,
  Neelakantan, Shyam, Sastry, Askell, Agarwal, Herbert-Voss, Krueger, Henighan,
  Child, Ramesh, Ziegler, Wu, Winter, Hesse, Chen, Sigler, Litwin, Gray, Chess,
  Clark, Berner, McCandlish, Radford, Sutskever, and Amodei}]{gpt3}
Tom Brown, Benjamin Mann, Nick Ryder, Melanie Subbiah, Jared~D Kaplan, Prafulla
  Dhariwal, Arvind Neelakantan, Pranav Shyam, Girish Sastry, Amanda Askell,
  Sandhini Agarwal, Ariel Herbert-Voss, Gretchen Krueger, Tom Henighan, Rewon
  Child, Aditya Ramesh, Daniel Ziegler, Jeffrey Wu, Clemens Winter, Chris
  Hesse, Mark Chen, Eric Sigler, Mateusz Litwin, Scott Gray, Benjamin Chess,
  Jack Clark, Christopher Berner, Sam McCandlish, Alec Radford, Ilya Sutskever,
  and Dario Amodei. 2020.
\newblock \href
  {https://proceedings.neurips.cc/paper/2020/file/1457c0d6bfcb4967418bfb8ac142f64a-Paper.pdf}
  {Language models are few-shot learners}.
\newblock In \emph{Advances in Neural Information Processing Systems},
  volume~33, pages 1877--1901. Curran Associates, Inc.

\bibitem[{Brown-Schmidt et~al.(2015)Brown-Schmidt, Yoon, and
  Ryskin}]{brown2015people}
Sarah Brown-Schmidt, Si~On Yoon, and Rachel~Anna Ryskin. 2015.
\newblock People as contexts in conversation.
\newblock In \emph{Psychology of learning and motivation}, volume~62, pages
  59--99. Elsevier.

\bibitem[{Clark and Schaefer(1989)}]{clark1989contributing}
Herbert~H. Clark and Edward~F. Schaefer. 1989.
\newblock Contributing to discourse.
\newblock \emph{Cognitive Science}, 13(2):259--294.

\bibitem[{Clark and Wilkes-Gibbs(1986)}]{clark1986referring}
Herbert~H. Clark and Deanna Wilkes-Gibbs. 1986.
\newblock Referring as a collaborative process.
\newblock \emph{Cognition}, 22(1):1--39.

\bibitem[{Connell and Lynott(2014)}]{ConnelLynott2014}
Louise Connell and Dermot Lynott. 2014.
\newblock Principles of representation: {W}hy you can't represent the same
  concept twice.
\newblock \emph{Topics in Cognitive Science}.

\bibitem[{DeJong(1982)}]{dejong1982overview}
Gerald DeJong. 1982.
\newblock An overview of the {FRUMP} system.
\newblock \emph{Strategies for {N}atural {L}anguage {P}rocessing},
  113:149--176.

\bibitem[{Edmundson(1969)}]{edmundson1969new}
Harold~P Edmundson. 1969.
\newblock New methods in automatic extracting.
\newblock \emph{Journal of the ACM (JACM)}, 16(2):264--285.

\bibitem[{Frank and Goodman(2012)}]{frank2012predicting}
Michael~C Frank and Noah~D Goodman. 2012.
\newblock Predicting pragmatic reasoning in language games.
\newblock \emph{Science}, 336(6084):998--998.

\bibitem[{Franke et~al.(2010)Franke, J{\"a}ger, and
  Rooij}]{franke2010vagueness}
Michael Franke, Gerhard J{\"a}ger, and Robert~van Rooij. 2010.
\newblock Vagueness, signaling and bounded rationality.
\newblock In \emph{JSAI international symposium on artificial intelligence},
  pages 45--59. Springer.

\bibitem[{Goel and Byrne(2000)}]{goel2000minimum}
Vaibhava Goel and William~J Byrne. 2000.
\newblock Minimum {B}ayes-risk automatic speech recognition.
\newblock \emph{Computer Speech and Language}, 14(2):115--135.

\bibitem[{Grice(1975)}]{grice1975logic}
Herbert~P Grice. 1975.
\newblock Logic and conversation.
\newblock In \emph{Speech acts}, pages 41--58. Brill.

\bibitem[{Haber et~al.(2019)Haber, Baumg{\"a}rtner, Takmaz, Gelderloos, Bruni,
  and Fern{\'a}ndez}]{haber2019photobook}
Janosch Haber, Tim Baumg{\"a}rtner, Ece Takmaz, Lieke Gelderloos, Elia Bruni,
  and Raquel Fern{\'a}ndez. 2019.
\newblock \href {https://www.aclweb.org/anthology/P19-1184.pdf} {The
  {PhotoBook} dataset: Building common ground through visually-grounded
  dialogue}.
\newblock In \emph{Proceedings of the 57th Annual Meeting of the Association
  for Computational Linguistics}, pages 1895--1910.

\bibitem[{Hawkins et~al.(2020)Hawkins, Frank, and
  Goodman}]{hawkins2020characterizing}
Robert~D Hawkins, Michael~C Frank, and Noah~D Goodman. 2020.
\newblock Characterizing the dynamics of learning in repeated reference games.
\newblock \emph{Cognitive science}, 44(6):e12845.

\bibitem[{Holtzman et~al.(2019)Holtzman, Buys, Du, Forbes, and
  Choi}]{holtzman2019curious}
Ari Holtzman, Jan Buys, Li~Du, Maxwell Forbes, and Yejin Choi. 2019.
\newblock {The Curious Case of Neural Text Degeneration}.
\newblock In \emph{International Conference on Learning Representations}.

\bibitem[{Jaeger and Tily(2011)}]{jaeger2011language}
T~Florian Jaeger and Harry Tily. 2011.
\newblock On language ‘utility’: {P}rocessing complexity and communicative
  efficiency.
\newblock \emph{Wiley Interdisciplinary Reviews: Cognitive Science},
  2(3):323--335.

\bibitem[{Krahmer and van
  Deemter(2012)}]{krahmer-van-deemter-2012-computational}
Emiel Krahmer and Kees van Deemter. 2012.
\newblock \href {https://doi.org/10.1162/COLI_a_00088} {Computational
  generation of referring expressions: {A} survey}.
\newblock \emph{Computational Linguistics}, 38(1):173--218.

\bibitem[{Krauss and Weinheimer(1964)}]{krauss1964changes}
Robert~M Krauss and Sidney Weinheimer. 1964.
\newblock Changes in reference phrases as a function of frequency of usage in
  social interaction: {A} preliminary study.
\newblock \emph{Psychonomic Science}, 1(1):113--114.

\bibitem[{Kumar and Byrne(2002)}]{kumar2002minimum}
Shankar Kumar and Bill Byrne. 2002.
\newblock Minimum {B}ayes-risk word alignments of bilingual texts.
\newblock In \emph{Proceedings of the 2002 Conference on Empirical Methods in
  Natural Language Processing (EMNLP 2002)}, pages 140--147.

\bibitem[{Kumar and Byrne(2004)}]{kumar2004minimum}
Shankar Kumar and William Byrne. 2004.
\newblock Minimum {B}ayes-risk decoding for statistical machine translation.
\newblock In \emph{Proceedings of the Human Language Technology Conference of
  the North American Chapter of the Association for Computational Linguistics:
  HLT-NAACL 2004}, pages 169--176.

\bibitem[{Luhn(1958)}]{luhn1958automatic}
Hans~Peter Luhn. 1958.
\newblock The automatic creation of literature abstracts.
\newblock \emph{IBM Journal of Research and Development}, 2(2):159--165.

\bibitem[{Premack and Woodruff(1978)}]{premack1978tom}
David Premack and Guy Woodruff. 1978.
\newblock Does the chimpanzee have a theory of mind?
\newblock \emph{Behavioral and Brain Sciences}, 1(4):515--526.

\bibitem[{Reddy(1977)}]{reddy-1977}
Raj Reddy. 1977.
\newblock \href {https://doi.org/10.1184/R1/6609821.v1} {Speech understanding
  systems: {A} summary of results of the five-year research effort at
  {C}arnegie {M}ellon {U}niversity}.

\bibitem[{Reiter and Dale(1997)}]{reiter1997building}
Ehud Reiter and Robert Dale. 1997.
\newblock Building applied natural language generation systems.
\newblock \emph{Natural Language Engineering}, 3(1):57--87.

\bibitem[{Shannon(1948)}]{shannon1948}
Claude~E. Shannon. 1948.
\newblock A mathematical theory of communication.
\newblock \emph{The Bell System Technical Journal}, 27(3):379--423.

\bibitem[{Shore et~al.(2018)Shore, Androulakaki, and
  Skantze}]{shore-etal-2018-kth}
Todd Shore, Theofronia Androulakaki, and Gabriel Skantze. 2018.
\newblock \href {https://aclanthology.org/L18-1123} {{KTH} tangrams: {A}
  dataset for research on alignment and conceptual pacts in task-oriented
  dialogue}.
\newblock In \emph{Proceedings of the Eleventh International Conference on
  Language Resources and Evaluation ({LREC} 2018)}, Miyazaki, Japan. European
  Language Resources Association (ELRA).

\bibitem[{Simon(1990)}]{simon1990bounded}
Herbert~A Simon. 1990.
\newblock Bounded rationality.
\newblock In \emph{Utility and probability}, pages 15--18. Springer.

\bibitem[{Sumers et~al.(2021)Sumers, Hawkins, Ho, and
  Griffiths}]{sumers2021extending}
Ted Sumers, Robert Hawkins, Mark~K Ho, and Tom Griffiths. 2021.
\newblock Extending rational models of communication from beliefs to actions.
\newblock In \emph{Proceedings of the Annual Meeting of the Cognitive Science
  Society}, volume~43.

\bibitem[{Tomasello(2005)}]{tomasello2005constructing}
Michael Tomasello. 2005.
\newblock \emph{Constructing a language: {A} usage-based theory of language
  acquisition}.
\newblock Harvard university press.

\bibitem[{Zaslavsky et~al.(2021)Zaslavsky, Hu, and
  Levy}]{zaslavsky-etal-2021-rate}
Noga Zaslavsky, Jennifer Hu, and Roger~P. Levy. 2021.
\newblock \href {https://aclanthology.org/2021.scil-1.32} {A
  {R}ate{--}{D}istortion view of human pragmatic reasoning?}
\newblock In \emph{Proceedings of the Society for Computation in Linguistics
  2021}, pages 347--348, Online. Association for Computational Linguistics.

\bibitem[{Zipf(1949)}]{zipf1949human}
G.K. Zipf. 1949.
\newblock \emph{Human Behavior and the Principle of Least Effort: {A}n
  Introduction to Human Ecology}.
\newblock Addison-Wesley Press.

\end{thebibliography}
\bibliographystyle{acl_natbib}

\end{document}